\documentclass[11pt,a4paper]{article}
\usepackage[dvipsnames]{xcolor}
\usepackage[hyperref]{acl2021}
\usepackage{times}
\usepackage{latexsym}

\usepackage{microtype}

\aclfinalcopy 

\usepackage{graphicx}
\usepackage{comment}

\title{Evaluation of NMT-Assisted Grammar Transfer for a Multi-Language Configurable Data-to-Text System}

\author{Andreas Madsack \and Johanna Heininger \and Adela Schneider\\
{\bf Ching-Yi Chen} \and {\bf Christian Eckard} \and {\bf Robert Wei{\ss}graeber} \\
AX Semantics, Stuttgart, Germany \\
{\tt \{firstname.lastname\}@ax-semantics.com}}

\date{}

\begin{document}
\maketitle
\begin{abstract}

One approach for multilingual data-to-text generation is to translate grammatical configurations upfront from the source language into each target language. These configurations are then used by a surface realizer and in document planning stages to generate output.

In this paper, we describe a rule-based NLG implementation of this approach where the configuration is translated by Neural Machine Translation (NMT) combined with a one-time human review, and introduce a cross-language grammar dependency model to create a multilingual NLG system that generates text from the source data, scaling the generation phase without a human in the loop.

Additionally, we introduce a method for human post-editing evaluation on the automatically translated text.
Our evaluation on the SportSett:Basketball dataset shows that our NLG system performs well, underlining its grammatical correctness in translation tasks.

\end{abstract}

\section{Introduction}

Deep learning-based approaches offer promising results in the field of multi-language data-to-text generation. Being prone to hallucinations, they require a manual review post generation. To address this challenge, we propose a hybrid approach that integrates the rule-based system with deep learning models to leverage the strengths of each while avoiding hallucination. Our motivation behind incorporating the rule-based approach is to ensure accurate multilingual output, since guaranteeing accuracy is the key to scaling multi-language data-to-text systems - avoiding the human in the loop for each produced text.

The essential component of our approach is the configurable grammar unit, which carries linguistic information and can be transferred across languages. In this paper, our primary goal is to evaluate the effectiveness of transferring these grammar units across languages during translation. Since grammar transfer is the key to effectively adapting our rule-based system \cite{weissgraeber-madsack-2017-working} to multi-language settings, it directly influences the accuracy of our generated texts, and therefore, is worth to be evaluated. 

We aim to answer two research questions: firstly, how can we effectively evaluate our data-to-text system which incorporates NMT with grammar transfer? To answer that, we developed an evaluation framework, where we track edits of human translators after automatic translation. Secondly, how good is our grammar transfer feature in this hybrid system? We answer it by analyzing what types of post-edits are needed after the automatic translation, which helps us to precisely identify the errors in our system, with a specific focus on the grammar transfer mechanism.


\section{System description} 
\subsection{Data-to-Text System}

A data-to-text NLG system receives as input (a) data and (b) a specification how to generate the resulting human language text.
Such a system can be used to automate journalistic texts, financial reports or product descriptions in e-commerce.
Correctness is essential for all of these kinds of texts. Therefore, controlling the grammar features to ensure a correct grammatical specification of the system is worthwhile.

The NLG generation process has three fundamental steps \citep{reiter/dale:2000}: Document Planning, Microplanning, and Surface Realisation.
Document Planning is about which data are actually part of the resulting text and its overall structure. Microplanning is about the grammatical specification needed to generate an error-free human language text in the third step, Surface Realisation.

In our system, the Document Planning stage is based on the sentences the user annotates: which data fields are used and which sentences are part of their story. As for grammatical specification in Microplanning, it is done by the user but with the assistance of our NLG platform, which provides only valid options for the target language in the user interface.
For example, the user can decide to use the number of goals from the data and construct the following sentence: "Peter scored [number of goals] [goal + number=number of goals]."

\textit{Configurability} -- in the Surface Realisation step, the grammatical specifications are also needed for the variables of the sentence to be in the correct grammatical agreement. Surface Realisation is a completely automatic step based on the given specifications and on extensive language knowledge. Examples of such systems are \citet{gatt-reiter-2009-simplenlg} for English, \citet{braun-etal-2019-simplenlg} for German, and commercial \citep{dale2020} and proprietary ones \citep{weissgraeber-madsack-2017-working}.

\subsection{Multilinguality} 


For data-to-text projects that require the generation of text in more than one language, there are various approaches available. One way to achieve text in multiple languages, is by translating the generated text of a source language to a target language, another way is translating the generation rules from source to target language - the latter being more common for rule-based systems.

In either case, \textbf{Neural Machine Translation} is the state-of-the-art for translating plain text. That is, a neural network based on a Transformer architecture \cite{vaswani2017attention} is used to translate a sequence of input tokens from the source language to a sequence of tokens in the target language in an end-to-end manner. Those networks are trained on a large set of text pairs and the trained model is then applied to each new translation task.

\subsection{Grammar Transfer}

\begin{figure*}
    \centering
    \includegraphics[width=.7\paperwidth]{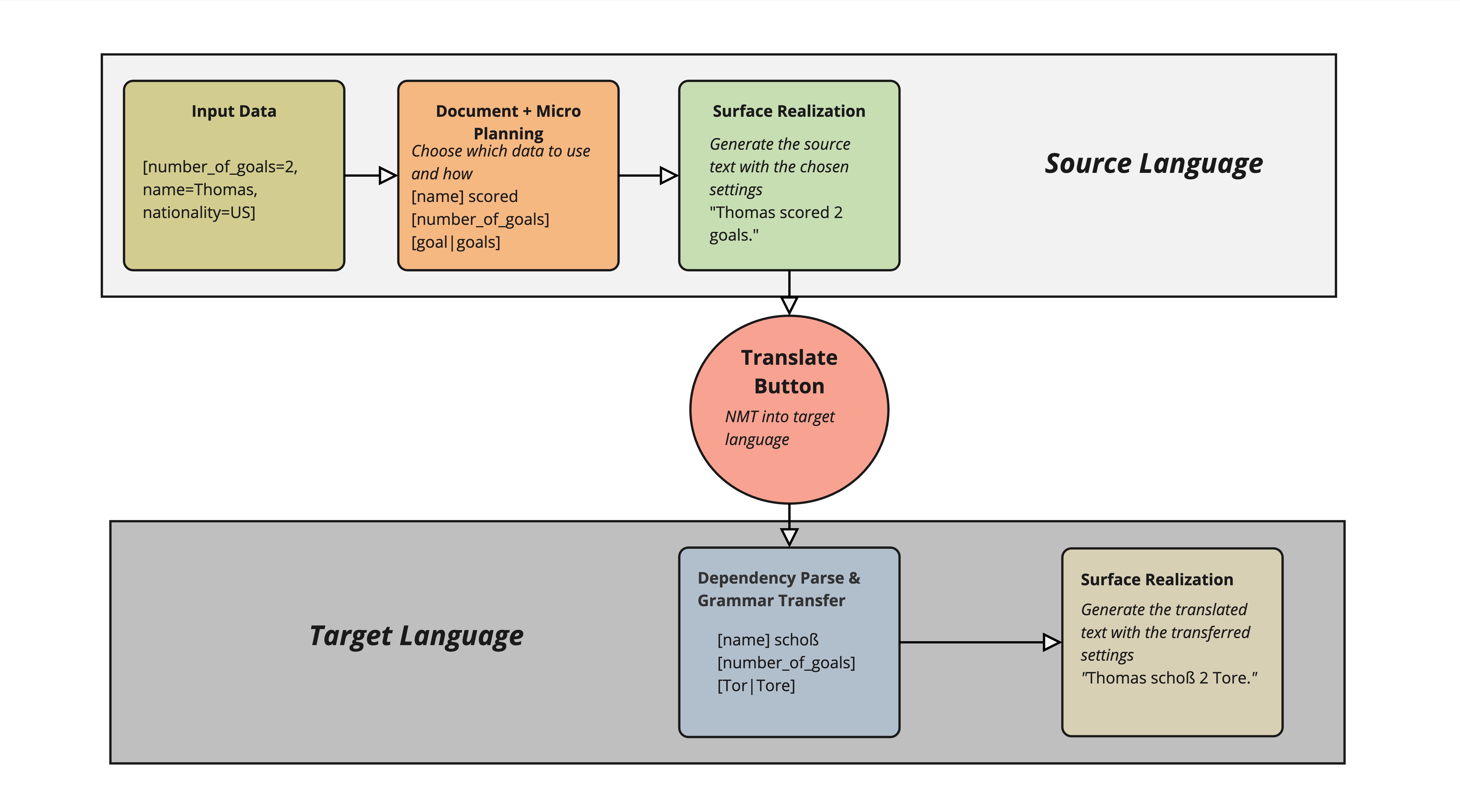}
    \caption{Translation workflow, which involves adapting grammatical units during text translation to ensure proper grammatical alignment between source and target languages.}
    \label{fig:translation_workflow}
\end{figure*}

When translating NLG projects on our platform, the grammatical specifications for the variable parts also need to be translated. Technically, we introduce the \textbf{grammar unit} as a container of grammar settings. Grammar units not only contain the grammar setting (e.g., number, case, tense, or possessor) but also can be transferred from one language to another language. 
To be able to do this, we use Natural Language Processing with the help of spaCy \citep{ines_montani_2023_7936855} on these grammar units. As Figure~\ref{fig:translation_workflow} shows,
after the source text was translated by NMT, each previously marked text snippet belonging to a grammar unit is analysed by SpaCy with the corresponding language-specific model and a dependency parse of the phrase is returned. Dependency parsing has been a common way to represent the syntactical structure in a sentence for quite some time \citep{Carroll1992TwoEO}.
  
We post-process the spaCy parse tree with custom aggregations of noun, pronoun, and verb grammar units.
Grammatical settings that are transferred can be seen in Table~\ref{table:feats}.

\begin{table}[ht]
\centering
{\footnotesize
\begin{tabular}{l|c|c|c}
\textbf{Features} & \textbf{Noun} & \textbf{Pronoun} & \textbf{Verb}\\
\hline
\hline
lemma     & x    &  x  & x \\
\hline
case      & x    &  x  & - \\
\hline
number    & x    &  x  & x \\
\hline
tense     & -    &  -  & x \\
\hline
person    & -    &  -  & x \\
\hline
gender    & x    &  x  & x \\
\hline
preposition    & x    &  x  & - \\
\hline
adjectives    & x    &  -  & - \\
\hline
numerals    & x    &  -  & - \\
\hline
conjunctions   & x    &  -  & - \\
\hline
determiners    & x    &  -  & - \\
\hline
pronoun type   & -    &  x  & - \\
\end{tabular}
}
\caption{Features per part of speech}
\label{table:feats}
\end{table}


Table~\ref{table:transfer} shows an example of transferring grammar units from English to German. Here the English noun phrase ``on Saturday'' with the preposition ``on'' and the noun lemma ``Saturday'', which is in nominative singular, is translated to the German noun phrase ``am Samstag''. It can be seen that, besides the literal translation, there are two transfers happening: 1) a definite determiner is added to the phrase and 2) the case changes to dative.


\begin{table}[ht]
\centering
{\footnotesize
\begin{tabular}{l|c|c|c|c}
\textbf{language} & \textbf{prep} & \textbf{det} & \textbf{noun} & \textbf{case}\\
\hline
\hline
English     & on    &  -         & Saturday & nominative \\
\hline
German      & an    &  definite  & Samstag  & dative \\
\end{tabular}
}
\caption{Transfer of the grammar unit ``on Saturday'' in English to ``am Samstag'' in German}
\label{table:transfer}
\end{table}


\section{Background}
The question of how to adequately evaluate generated texts is currently the subject of much discussion for text generation \cite{howcroft-etal-2020-twenty, gehrmann-etal-2021-gem} as well as machine translation \cite{freitag-etal-2021-experts}. 
 The focus is on developing standardized metrics for automated evaluation and clarifying the requirements for human evaluation.  In the discussion of the latter, the need to clearly define the quality criteria used is emphasized (\citet{belz-etal-2020-disentangling}).  

\textit{Grammaticality} -- the quality criterion used in our study -- falls under the more general correctness of surface form criterion \citep{gehrmann2022repairing} or is subsumed under larger concepts (for \textit{fluency} \citealp{pan-etal-2020-semantic}
\textit{coherence} \citealp{juraska-etal-2019-viggo}). 
Omitting such a larger scope for the quality criterion also leads to the choice of the evaluation method: The goal is to identify the errors rather than to asses a broader concept of the text quality. 

Methods for error identification such as annotation by (professional) translators, post-editing, or error class analysis are more common in machine translation than in NLG \citep{avramidis-etal-2012-involving}.
Particularly post-editing plays an important role there: On the one side as a practice in the translator's work, partly built into MT systems, and on the other side also as a human evaluation method \citep{aziz-etal-2012-pet}. 

In the field of NLG this evaluation method is used only sporadically \citep{howcroft-etal-2020-twenty}. Such an individual case is the evaluation of a data-to-text system which is generating weather reports by \citet{sripada-etal-2005-evaluation, sripada:2004}, whereas here the results were edited one by one. One reason the authors give for choosing this method is that it can provide insight into the usefulness of the system for real users. This aspect is also relevant to us. 

When we ask for a post-edit, the participants are supposed to take their grammatical acceptability \citep{t2016empirical} as the basis for the edit. At the same time, the system gives them specific constraints and limited options: They need to think about what they have judged in formal grammatical terms and specify this in the system. 

\section{Experiment}
\subsection{Dataset}
We use the Sport-Sett:Basketball dataset\footnote{\href{https://huggingface.co/datasets/GEM/sportsett_basketball}{https://huggingface.co/datasets/GEM/sportsett\_basketball}} \cite{thomson-etal-2020-sportsett} without the summary reference from its corpus. Instead, we use instances of this dataset to generate new texts as references that contain more grammatical features, see Table~\ref{table:feats}.

In the experiment, source text generation and target text translation take place in our proprietary platform. There, we perform Document Planning, Microplanning, Surface Realization, Machine Translation as well as human review.




\subsection{Post-edit experiment} 
The post-edit experiment aims to measure the number of post-edits as well as the category of post-edits that a translator needs to perform to optimize the system output. Our focus is particularly on translation errors in grammar. There are 13 translators\footnote{Specifically, there are seven translators for German, one for French, one for Spanish, one for Slovenian, one for Polish, one for Portuguese, and one for Chinese.} in this human evaluation, and they are native speakers or have a native-speaker level. English is the source language for all translators. 

Our experiment asks translators to review and revise translations at the statement level, where each statement may contain more than one sentence. Grammar features shown in Table~\ref{table:feats} are explicitly aligned with a corresponding source grammar. Both source and target statements are displayed horizontally next to each other for review. 

Revising previous statements is allowed. In addition, each source and target statement pair can be generated in four versions with different input data. This enables translators to judge grammar from multiple perspectives. The translators were asked to identify all errors and adjust the settings or text of the grammar units, such that the newly generated text is correct. Every post-editing has been tracked in the text and tagged with corresponding change categories in Table~\ref{table:changes_per_grammar_unit}.


\section{Results}
We evaluated the changes between the automatic process artifacts (NMT combined with grammar transfer) and the human post-editing with focus on changes made to the grammar units and not the rendered text.
Matching of the grammar units happened semi-automatically.
A unit may be removed by the annotator or new units may be created.
We were able to match 98\% of the grammar units between automatic translation and human post-edit translation.

Table~\ref{table:changes_per_grammar_unit} shows the changes per grammar unit as percentage of all grammar units of the language. Each change helps us to evaluate and precisely identify the errors in our system.
The most common change for German was ``change case'' of a noun phrase with 8\% of all the German grammar units in the experiments.
The actual change here was mostly from ``genitive'' to ``nominative'' case because the dependency parser classified names of the basketball teams (e.g., ``Denver Nuggets'') as singular genitive because they end with an ``s''.
The second most common change for German is marking the head of a compound noun to fix missing lexical information. For example, marking ``Ergebnis'' as the head of ``Double-Double-Ergebnis'' improves German inflection correctness.

\begin{table}[ht]
{\scriptsize
\begin{tabular}{l|c|c|c|c|c|c|c}
language & \rotatebox[origin=c]{90}{de-DE} & \rotatebox[origin=c]{90}{es-ES} & \rotatebox[origin=c]{90}{fr-FR} & \rotatebox[origin=c]{90}{pl-PL} & \rotatebox[origin=c]{90}{pt-BR} & \rotatebox[origin=c]{90}{sl-SI} & \rotatebox[origin=c]{90}{zh-CN} \\
change (\%) &  &  &  &  &  &  &  \\
\hline
\hline
add adjective & 1 &  &  &  &  &  &  \\
add determiner & 1 & 1 &  &  &  &  &  \\
add noun & 1 &  &  &  &  &  &  \\
add number & 1 & 1 &  &  &  &  &  \\
add preposition & 1 &  & 1 &  &  &  & 7 \\
add pronoun &  &  &  &  &  &  & 1 \\
capitalize &  & 1 &  &  &  & 1 &  \\
change POS & 1 &  &  &  &  &  &  \\
change adjective lemma & 2 & 1 & 2 &  &  & 1 &  \\
change case & 7 &  &  & 1 &  & 8 &  \\
change conjunction & 1 & 1 &  &  &  & 1 & 4 \\
change determiner & 1 &  & 1 &  &  &  &  \\
change noun lemma & 2 & 5 & 15 & 3 & 3 & 5 & 19 \\
change number & 3 & 3 &  &  &  & 1 &  \\
change numeral type & 1 & 1 &  & 1 &  & 1 &  \\
change preposition &  & 1 &  &  & 5 &  &  \\
change tense & 1 & 1 & 2 &  &  & 1 &  \\
change verb lemma & 1 & 2 & 1 &  &  &  &  \\
lowercase & 1 &  &  &  &  & 6 &  \\
mark head & 5 & 1 &  &  & 1 &  &  \\
remove adjective & 1 &  &  &  &  &  & 1 \\
remove determiner & 1 & 1 &  &  &  &  &  \\
remove preposition & 1 & 1 &  &  & 1 &  &  \\
\end{tabular}
}
\caption{Percentage of changes per grammar unit for every language}
\label{table:changes_per_grammar_unit}
\end{table}

Changes in German, Spanish, French, Portuguese, Slovenian, and Chinese were as expected and led to a correct surface rendering of the grammar units, but in Polish the results were incomplete and the translator has not finished the experiment. 

On average only \textbf{19\%} of the grammar units were actually changed, so most of them were correctly translated and transferred.

\begin{figure}[ht]
\includegraphics[scale=.48]{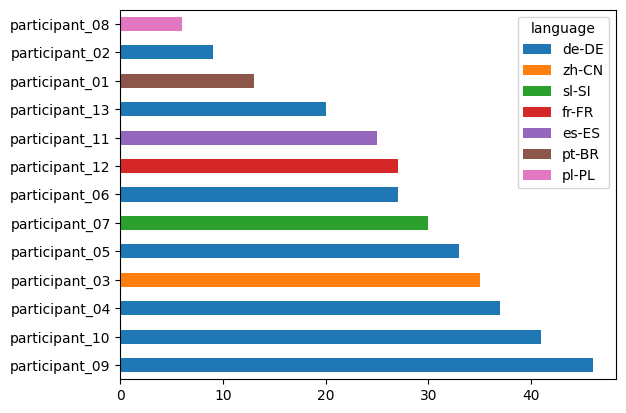}
\caption{Number of changes per participant}
\label{fig:num_changes}
\end{figure}

This can be seen in Figure~\ref{fig:num_changes} where  participant no.8 is the one with the fewest changes.
One other observation is the amount of changes per participant can vary a lot even within a language. Participant no.2 did the absolute minimum of changes to fix a German translation and not everything is absolutely perfect, but grammatically correct. On the contrary, participant no.9 did the most changes in German, even adding more grammar units that were not necessary to get a grammatically correct rendered text. We hope to better measure this bandwidth of changes by having more participants in the next iteration of the study.

\section{Conclusion and Future Work}
We have shown that NMT can be integrated into the workflow of rule-based data-to-text systems. With the additional benefit in terms of speed and linguistic quality of NMT that can be utilized without losing the advantage of the configurability.

In order to guarantee grammatical correctness, it is essential that the grammatical specifications of the dynamic elements can be transferred from the source language to the target language without any errors. 

To test the quality of this grammar transfer, we evaluated the post-editing actions of the translators. This showed that the automatic transfer was error-free in most cases. However, this high rate is not sufficient for an uncontrolled output of the translation for a real-world application. There the correction of grammar units by human translators remains essential. This effort however is small compared to the individual checking of large amounts of generated and translated texts.

A brief analysis of the errors indicates that it is possible to improve the automatic transfer actions at least for certain types of errors.

Although the post-editing method has provided sufficient results for our goals by providing qualitative statements about errors and their correction as well as an assessment of the system's usability, we are nevertheless aware of the limitations of the evaluation: The limited number of participants hardly allows a profound quantitative evaluation. In addition, the reproducibility is limited because we used a closed source platform. 

Therefore, we plan for our future work to conduct an extensive study with more translators. We are also interested in investigating translation performance when using grammar-rich languages such as German as a source language. 





\bibliographystyle{acl_natbib}
\bibliography{acl2021}

{\onecolumn
\appendix
\section{Appendix}

\begin{table}[ht]
\centering
\begin{tabular}{|p{4.9cm}|p{4.9cm}|p{4.9cm}|}
\hline
\multicolumn{3}{|c|}{Data instance: 97 (Bulls-Nuggets)}\\
\hline
English source text &German translation by NMT &German translation by NMT \\ & &+ human post-edit \\
\hline
\textbf{Chicago Bulls vs.\ Denver Nuggets} &\textbf{Chicago Bulls vs.\ Denver Nuggets} &\textbf{Chicago Bulls vs.\ Denver Nuggets} \\
The match between the Chicago Bulls and the Denver Nuggets took place on Thursday (January 1, 2015) at the sold-out United Center (Illinois). &Das Spiel zwischen Chicago Bulls und Denver Nuggets fand am DonnerstagJanuar 1, 2015) im vollständigen United CenterIllinois) statt. &Das Spiel zwischen \textcolor{Green}{den} Chicago Bulls und \textcolor{Green}{den} Denver Nuggets fand am Donnerstag \textcolor{Green}{(01. Januar} 2015) im vollständig gefüllten United Center \textcolor{Green}{(}Illinois) statt. \\
\hline
Over 20000 enthusiastic fans came to Chicago on the 1st gameday of the 2015 season, completely filling the stadium.  & Über 20000 leidenschaftliche Fans kamen am 1.\ Spieltag der Saison 2015 nach Chicago und füllten das Stadion komplett. & Über 20000 leidenschaftliche Fans kamen am 1. Spieltag der Saison 2015 nach Chicago und füllten das Stadion komplett.  \\
\hline
The home team won with 106 - 101 against the visiting team from Denver. & Die Heimmannschaft gewann mit 106 - 101 gegen die Gastmannschaft aus Denver. & Die Heimmannschaft gewann mit 106\textcolor{Green}{:}101 gegen die Gastmannschaft aus Denver. \\
\hline
The best player of the game was undoubtedly Jimmy Butler of the Chicago Bulls. He led the team to victory with 26 points, 8 impressive rebounds, 8 precise assists, 2 crucial steals, and 1 spectacular block. & Der beste Spieler des Spiels war zweifelsohne Jimmy Butler von Chicago Bulls. Er führte das Team mit 26 Punkten, 8 Rückprall, 8 helfen, 2 stehlen und einem 1 Spektakulären Block zum Sieg. & Der beste Spieler des Spiels war zweifelsohne Jimmy Butler von \textcolor{Green}{den} Chicago Bulls. Er führte das Team mit 26 Punkten, \textcolor{Green}{8 beeindruckenden Rückprallern}, 8 \textcolor{Green}{geschickten Vorlagen}, 2 \textcolor{Green}{spielverändernden Steals} und einem 1 \textcolor{Green}{s}pektakulären Block zum Sieg.  \\
\hline
On the contrary, the Denver Nuggets were unable to secure a win despite the impressive performances of their top player.\ Wilson Chandler was the leading scorer on the team with 22 points... & Im Gegenteil, Denver Nuggets konnten trotz der beeindruckenden Leistungen ihres Spitzenspielers keinen Sieg erringen. Wilson Chandler war mit 22 Punkten der Topscorer des Teams... & Im Gegenteil, \textcolor{Green}{die} Denver Nuggets konnten trotz der beeindruckenden Leistungen ihres Spitzenspielers keinen Sieg erringen. Wilson Chandler war mit 22 Punkten der Topscorer des Teams...  \\
\hline
\end{tabular}
\caption{\label{tab:text_example}The source text, auto-translated text, as well as the auto-translated with human (participant 09) post-edit text from our system, using the Sport-Sett:Basketball dataset \cite{thomson-etal-2020-sportsett}. The green texts indicate the changes after the participant's post-edits. }
\end{table}}

\end{document}